\documentclass[11pt,letterpaper]{article}
\usepackage[letterpaper]{geometry}
\usepackage{acl2012}
\usepackage{times}
\usepackage{latexsym}
\usepackage{amsmath}
\usepackage{amssymb}
\usepackage{multirow}
\usepackage{url}
\usepackage[table]{xcolor}
\usepackage{underscore}
\usepackage{tikz}
\usepackage{ifthen}
\usepackage{pgfplots}
\usepackage{soul}

\makeatletter
\newcommand{\@BIBLABEL}{\@emptybiblabel}
\newcommand{\@emptybiblabel}[1]{}
\makeatother

\pgfplotsset{compat=1.10}
\usetikzlibrary{bayesnet}
\definecolor{leaRed}{rgb}{0.722, 0, 0.278}
\definecolor{leaGreen}{rgb}{0.13, 0.67, 0.8}

\title{Prosodic Features from Large Corpora of Child-Directed Speech \\as Predictors of the Age of Acquisition of Words}

\author{Lea Frermann \\
  University of Edinburgh \\
  {\tt  l.frermann@ed.ac.uk } \\\And
  Michael C. Frank \\
  Stanford University \\
  {\tt mcfrank@stanford.edu}}

\date{}
  
\begin{document}
\maketitle

\begin{abstract}
The impressive ability of children to acquire language is a widely studied phenomenon, and the factors influencing the pace and patterns of word learning remains a subject of active research. Although many models predicting the age of acquisition of words have been proposed, little emphasis has been directed to the raw input children achieve. In this work we present a comparatively large-scale multi-modal corpus of prosody-text aligned child directed speech. Our corpus contains automatically extracted word-level prosodic features, and we investigate the utility of this information as predictors of age of acquisition. We show that prosody features boost predictive power in a regularized regression, and demonstrate their utility in the context of a multi-modal factorized language models trained and tested on child-directed speech.
\end{abstract}

\section{Introduction}
Out of the many challenges infants face throughout their first years, the task of learning words of a language, i.e., mapping seemingly arbitrary sounds to concepts in the world, is particularly stunning. How do children accomplish this, and what factors influence the characteristics of this process?

This work aims to shed light on these question, approaching it from the perspective of predicting the age of acquisition of words. Diverse research efforts have addressed this question in recent years by collecting data sets revealing the time course in which language is acquired~\cite{Kuperman:2012,Frank:2016}, or investigating the influence of various linguistic (e.g., word frequency or mean length of utterance) and semantic (e.g., babiness or concreteness) on word age of acquisition~\cite{Braginsky:2016,Gentner:2001,Snedeker:2007}.

Past research efforts have been shaped by two major shortcomings. The first one relates to the reference data of age of acquisition of words which traditionally has consisted of adult estimates of the age at which they learnt a particular word (e.g.,~\newcite{Kuperman:2012}). Clearly these estimates are highly influenced by subjective inaccuracies. A more recent approach~\cite{Braginsky:2016} has used data from the MacArthur-Bates Communicative Development Index (CDI; \newcite{Fenson:2007}), which comprises large sets of parent-filled questionnaires of the words their child understands and/or produces at a particular age~\cite{Fenson:2007,Frank:2016}. Such questionnaires exist for a large number of children of a variety of mother tongues, and consequently provide a reliable and reasonable scale data set for predictive models. We will evaluate our models against this resource in this work.

The second shortcoming relates to the kind of predictors of age of acquisition which have been investigated in the past. Recent work \cite{Braginsky:2016} has investigated a variety of predictors including input-derived properties such as word frequency or mean length of utterance, as well as semantic properties such as concreteness and babiness. None of these predictors, however, directly consider the full characteristics of child-directed speech the child receives.

% This work scales up prior work on age of acquisition research in two ways. First, we utilize large amounts of child-directed speech from the CHILDES corpus~\cite{MacWhinney:2000}. Computational models of word learning~\cite{Fazly:2010} have attempted to predict age of acquisition as part of their output. Such models, however, rely on artificial 'scenes' of object which are typically constructed ad-hoc (by pairing words with symbolic meaning representations and adding random noise), and 

Here we advance prior work by investigating the impact of raw language input on age of acquisition.\footnote{Supplementary material (code and data) is available at \url{https://github.com/ColiLea/prosodyAOA}} In particular we are interested in the {\it prosodic} properties of individual words in child-directed speech. We derive generic sets of prosodic features (eGemaps features; \newcite{Eyben:2016}) information from large amounts of raw child-directed data, and investigate their utility in predicting the age of acquisition of words. We also incorporate such features into a language model, which we view as a simplistic model of word learning, and in turn investigate the impact of prosodic features in this framework. We summarize our contributions below.
\begin{itemize}
 \item We construct a corpus of word-level aligned text-speech data of two portions of the English CHILDES corpus. From this corpus, we extract sets of word-type level prosodic eGemaps features. We present a pipeline utilizing open source toolkits and established prosodic feature sets. 
 \item We examine the power of this multi-modal corpus of raw child-directed speech as a predictor of age-of-acquisition, and evaluate our predictors against a recent large-scale gold standard of CDI parent questionnaire.
 \item We incorporate the prosodic eGemaps features into language models and show that prosodic information can reduce perplexity of words in the context of the CHILDES corpus.
 \item In contrast to previous work we (a) examine predictive power on different sub-corpora of CHILDES and (b) adopt a cross-validation paradigm as well as regularized (Ridge) regression in an attempt to avoid overfitting our regression models and ensure generalizability of our results.
\end{itemize}

\section{A Corpus of Word-Level Prosody Features in Child-directed Speech}
Even though the importance of multi-modal experience on language acquisition is fairly uncontroversial, relevant models have been largely trained and tested on text transcriptions of child-directed speech from the CHILDES corpus. Some exceptions include attempts to incorporate visual or pragmatic information in the input~\cite{Frank:2009,Yu:2005}. This information is typically derived from manual annotations of videos accompanying some of the CHILDES data, is inherently expensive to obtain, resulting in small data sets.

A large portion of the English CHILDES corpus, however, consists of the raw audio records of parent-child interactions together with their orthographic transcripts, as well as utterance-level text-audio alignments. From this data we derive word- and phone-level alignments and use this corpus to extract word-level prosodic characteristics (cf. Section \ref{egemaps-features} for more details).

Given recent advances in automatic speech recognition and downstream tasks like forced alignment, the aligned orthographic-audio CHILDES data provide a fruitful resource for large-scale investigations of the influence and characteristics of prosody in child-directed speech. \newcite{Elsner:2017} describe a similar approach to ours towards aligning text-audio child-directed data. Their approach was developed independently of and concurrently to our work.

We align and extract features for two CHILDES corpora. First, we use the Brent corpus~\cite{Brent:2001} which includes child-directed speech to 16 infants aged between 9 and 15 months ($\sim$~154,700 child-directed utterances). In addition, we use the the Providence corpus~\cite{Demuth:2006} which comprises child-directed speech to six children roughly between age 1;00 and 4;00 ($\sim$~259,800 child-directed utterances).

For each child in our corpora we consider child-directed language by its main interactor (always the mother) and remove any utterance whose orthographic transcription included noise markers.

\subsection{Word-level text-audio alignment of English CHILDES data}
\label{word-alignment}

\begin{figure*}
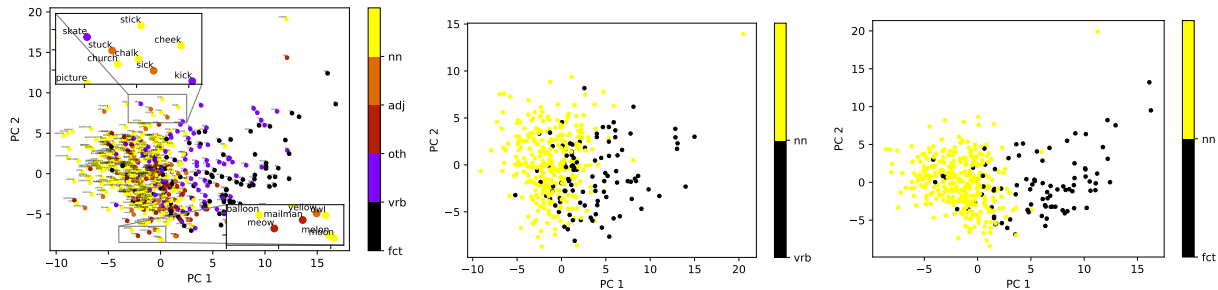

 \hspace{-6ex}
 \includegraphics[width=0.39\textwidth]{\detokenize{./BrentProvidence_full_zoom}}
 \includegraphics[width=0.32\textwidth]{\detokenize{./BrentProvidence_pca_nouns_verbs}}
 \includegraphics[width=0.32\textwidth]{\detokenize{./BrentProvidence_pca_nouns_function_words}}
\caption{Visualization of word clusters in vector space after projecting our 88-dimensional eGemaps feature representation into 2-dimensional space using principle component analysis (PCA). We color words based on their part of speech (nouns (nn), verbs (vrb), function words (fct), adjectives (adj), other (oth)).
\label{pca}}
\end{figure*}

We use the Montreal Forced Alignment (MFA) tool~\cite{McAuliffe:2017} to align orthographically transcribed child-directed speech to the corresponding audio snippets on the word level. The MFA tool also provides us with phoneme-level alignments which we have not made use of in this work. MFA uses Kaldi, a state-of-the-art forced aligner~\cite{Povey:2011}. Crucially, MFA takes into account speaker information and optimizes its alignments for individual speakers. Our corpus consisting of individual mother-child dyads is thus predestined for this approach.

MFA takes as input pairs of text and audio (wav) files containing the same information in orthographic and auditory modality, respectively. We provide one such pair for each child-directed utterance (taking utterance-level alignments as provided in the CHILDES corpus). MFA returns a {\it textgrid} file for each input pair, containing start and end time stamps for each word in the utterance, as well as for each phoneme in each word.

\subsection{Prosodic Feature extraction}

\label{egemaps-features}
We are now in a position where we can extract prosodic features of individual spoken words. \newcite{Eyben:2016} recently presented sets of core prosody features. Features were selected based on their empirical and theoretical value in prior work, and robust automatic computability, and include standard features based on $F0$ -- $F3$ formants, spectral shape and rhythm, intensity and MFCC features among others. We use their extended feature set which comprises a total of 88 features. For feature extraction, we use the scripts provided by the authors implemented within the OpenSMILE toolkit~\cite{Eyben:2013}.

Their scripts take as input the textgrid files produced by the MFA forced aligner, as well as the corresponding word-level audio snippets. They return an 88-dimensional feature vector for each spoken word in our corpus.

\subsection{Evaluation}
In order to estimate the quality of the automatically extracted eGemaps features, we test their discriminative power on established linguistic classes (part of speech). Concretely, we map the 88-dimensional eGemap feature vectors for 600 target words into 2-dimensional space using principle component analysis (PCA). We then inspect the extent to which words are clustered by their part of speech. We derive word-type level eGemaps representation from the word-token level eGemaps in our corpus by averaging all token-level features for each type. We combine the data from the Brent and Providence corpus for this analysis.

Figure~\ref{pca} displays the resulting space for various subsets of part of speech classes, which are color-coded. All dots in space correspond to a target word from the CDI data set, and are labeled with their word. It demonstrates that the transformed features are able to distinguish nouns from function words (right) and to some degree from verbs (centre). We also show clusters over all words in the CDI AoA data set (left), where nouns (yellow), verbs (violet) and function words (black) are well-separated along the first principle component. 

Additionally we observe that phonologically similar words are clustered together. For example the words {\it stick, cheek, chalk, sick, kick} all appear in a cluster in the top centre; and words {\it melon, meow, mailman, moon} appear in the bottom center of the space. We consider these results as initial support of the quality of our prosodic eGemaps features of child-directed speech.

\section{Experiment 1: Prosodic Features as Predictors of Age of Acquisition}
\begin{figure*}[ht]
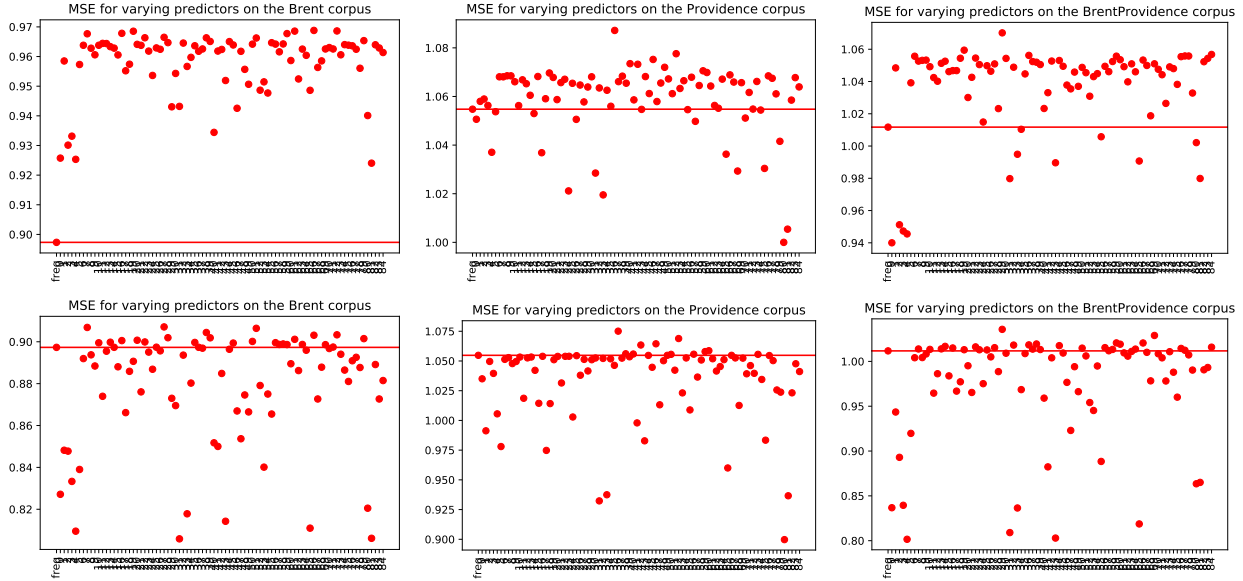

 \includegraphics[width=0.33\textwidth]{\detokenize{./Brent_egemaps_features_all}}
 \includegraphics[width=0.33\textwidth]{\detokenize{./Providence_egemaps_features_all}}
 \includegraphics[width=0.33\textwidth]{\detokenize{./BrentProvidence_egemaps_features_all}}
 \includegraphics[width=0.33\textwidth]{\detokenize{./Brent_egemaps_features_all_with_freq}}
 \includegraphics[width=0.33\textwidth]{\detokenize{./Providence_egemaps_features_all_with_freq}}
 \includegraphics[width=0.33\textwidth]{\detokenize{./BrentProvidence_egemaps_features_all_with_freq}}
 \caption{Ridge regression Mean Squared Error (MSE) of the frequency predictor and all individual eGemaps feature predictors on our three corpora Brent (left), Providence (centre), both (right). {\bf Top:} MSE for {\bf individual predictors}. {\bf Bottom: } MSE for {\bf frequency + eGemaps predictor}. The vertical line shows the performance of frequency as the sole predictor of age of acquisition.}
 \label{exp-1-all}
\end{figure*}
In this experiment we test the word-level eGemaps features as predictors of age of acquisition. We fit regularized ridge regression models including various subsets. Ridge regression is an extension of standard linear regression, which augments the sum of squares objective with a regularization term on the magnitude of the coefficients. This (a)~prevents the model from fitting the data too closely and generalizing poorly to new data; and (b)~to some extent alleviates the problem of feature collinearity.\footnote{We investigated collinearity between eGemaps features and observed strong correlations only between few selected feature pairs.}

We evaluate our models using 10-fold cross validation with random 90\% train and 10\% test splits. Note that this method departs from approaches in previous work~\cite{Braginsky:2016} where models were fitted to and tested on the same data set. We emphasize the importance of exploring any model's generalization ability by testing models on unseen data.

\subsection{Experiments and Results}
We fit ridge regression models to predict the age of acquisition of 600 target words in months.  We obtain word type level eGemap feature representations by averaging feature values across all (token) observations of a target word in our corpus. In addition we used word frequency as a predictor, which has been shown to be a strong predictor of age of acquisition in previous work~\cite{Braginsky:2016}. We scale all predictors to zero mean and unit variance and map frequency into log space.\footnote{Many of the eGemaps features are not strictly positive so we cannot map them to log space, and refrain from any other transformation for the time being.} Scores are reported as mean squared error (MSE) on the test set, so lower is better.

\subsubsection{Exploratory Egemaps Feature Analysis}
In this set of experiment we explore the 88 individual eGemaps features as well as word frequency as predictors of AoA.

Figure~\ref{exp-1-all} shows our results. We report results on our three corpora (Brent, Providence, both). The top row in Figure~\ref{exp-1-all} displays the mean squared error achieved by each individual eGemaps prosody feature as a predictor, plus the performance of word frequency as a predictor (leftmost data point, and emphasized by the horizontal red lines). For the Brent corpus frequency predicts AoA better than any prosodic feature, which is an expected result given the strong performance of frequency in prior work of AoA prediction~\cite{Braginsky:2016}. For the Providence (and consequently the combined) corpus, however, frequency is a less strong predictor than for the Brent corpus and is outperformed by selected prosodic features.

The bottom row of Figure~\ref{exp-1-all} shows MSE performance of the frequency predictor in combination with any individual eGemaps prosody feature. We can observe that a large number of prosodic features enhance prediction, i.e., explain variance above and beyond pure frequency-based models.

\subsubsection{Egemaps Feature Selection}
\begin{table}[t!]
\begin{small}
  \begin{tabular}{ll}
  \hline
  {\bf ID}   & {\bf Feature name}\\\hline
     p1 & F0semitoneFrom27.5Hz_sma3nz_amean\\
     p2 & F0semitoneFrom27.5Hz_sma3nz_percentile50.0\\
     p3 & F0semitoneFrom27.5Hz_sma3nz_percentile80.0\\
     p4 & F0semitoneFrom27.5Hz_sma3nz_stddevNorm\\
     p5 & hammarbergIndexUV_sma3nz_amean\\
     p6 & loudness_sma3_stddevRisingSlope\\
     p7 & * F0semitoneFrom27.5Hz_sma3nz_pctlrange0-2 \\
     p8 & * F1bandwidth_sma3nz_stddevNorm \\
     p9 & * jitterLocal_sma3nz_stddevNorm \\
     p10 & * MeanVoicedSegmentLengthSec \\
     p11 & * shimmerLocaldB_sma3nz_stddevNorm \\
     p12 & * spectralFluxV_sma3nz_stddevNorm \\
     p13 & * VoicedSegmentsPerSec    \\\hline
  \end{tabular}
  \end{small}
  \caption{Best performing Egemaps features on the Brent and Providence corpus as selected based on the exploratory feature analysis. Features marked with an asterisk (*) were among the 10 best performing features for both corpora.}
  \label{selected-features}
\end{table}

Based on our exploratory study, for both the Brent corpus and the Providence corpus we select the 10 eGemaps features which combined with the frequency predictor lead to best AoA prediction performance in terms of MSE. Out of these 20 features seven occur in both corpus-specific feature list, so that we obtain a final set of 13 individual features. The features are listed in Table~\ref{selected-features}.

\begin{figure*}[ht]
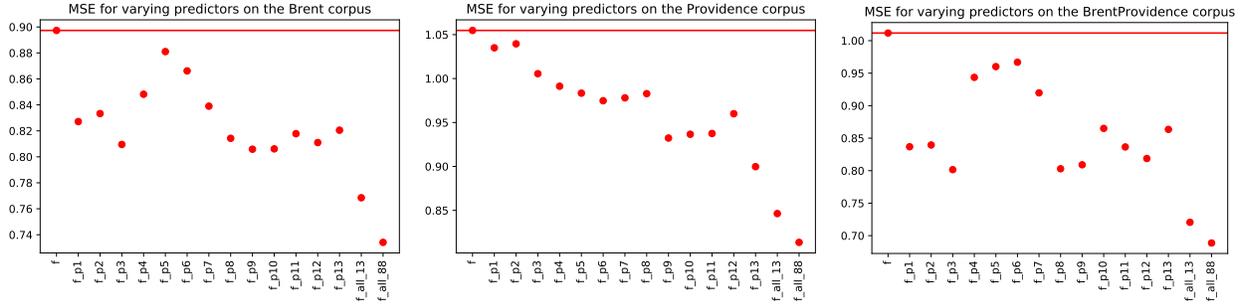

 \includegraphics[width=0.33\textwidth]{\detokenize{./Brent_egemaps_features_top10_with_freq}}
 \includegraphics[width=0.33\textwidth]{\detokenize{./Providence_egemaps_features_top10_with_freq}}
 \includegraphics[width=0.33\textwidth]{\detokenize{./BrentProvidence_egemaps_features_top10_with_freq}}
 \caption{Mean Squared Error (MSE) of the 13 {\bf selected} eGemaps feature predictors on our three training corpora (Brent, Providence, both). Table~\ref{selected-features} maps feature IDs to the actual prosodic feature.}
 \label{exp-1-selected}
\end{figure*}

Figure~\ref{exp-1-selected} shows mean squared error results of ridge regression models incorporating frequency (f) (leftmost data point, and emphasized through horizontal red bar) and one of our selected eGemaps features (p1--p13; cf. Table~\ref{selected-features} for details) across the three corpora, as well as all features combined. We see that all eGemaps based predictors lower MSE compared to using frequency only. We furthermore observe that combining all 13 eGemaps features and frequency (f_all_13) further improves prediction. Finally we combine frequency with our full set of 88 eGemaps features (f_all_88) which, as expected, results in best MSE. These results are stable across corpora.

\subsection{Discussion}
This experiment investigated the utility of automatically extracted word level prosodic features from child-directed speech as predictors of age of acquisition of words. We utilized a generic set of 88 prosodic features which have been previously proposed in the literature~\cite{Eyben:2016}.

We conducted an exploratory analysis of all 88 features as predictors of age of acquisition, and selected a the ten predictors which most notably boosted predictive performance over using word frequency as the sole predictor independently for our two training corpora. The fact that seven out of the 20 predictors (35\%) overlapped across the two corpora suggests that the selected feature subset is robust and generalizable across other data sets and tasks.

Regression models with large sets of potentially noisy predictors, like our automatically obtained prosody features, are inherently susceptible to discovering spurious regularities in the data set and overfit the training data without making meaningful predictions on unseen data. We address this issue by conducting all regression experiments within the framework of cross validation, i.e., random splits of the full training data set into train and test sets. We furthermore evaluate all our models on two data sets of child-directed speech, which can be viewed as two independent samples from the overall distribution of child-directed data. Overall we observe similar patterns across the corpora, further supporting the generalizability of our results.

While we could show that eGemaps features explain some of the variance of age of acquisition estimates beyond what is explained by frequency, the way eGemaps features were used implies a significant information loss. In particular, we averaged feature values across all mentions of any predicted word throughout the whole corpus.

Ideally, we would like to make use of each individual word mentions prosody features and their prosodic contexts. To this end, in the next set of experiments, we incorporate the eGemaps features into a (highly simplified) model of word learning. The model estimates word probabilities based on word mentions in local linguistic context in a large corpus of child directed speech. We define the context such that it encodes both lexical and prosodic information.

\section{Experiment 2: Enriching Language Models with Prosody Features}
\label{exp2}
In addition to employing prosody features directly as predictors of age of acquisition, we are interested in using such features in statistical models of word learning. As a very simple first attempt we investigate a classical n-gram language model as a highly constrained model of word learning. N-gram language models can be viewed as a purely syntactic model of word acquisition. The model is exposed to natural language sentences and learns conditional probability distributions over words given the immediate preceding context.

\subsection{Factored Language Models}
In a classical n-gram model this context is represented as the $n$ preceding words in a sentence. There have been numerous attempts, however, to enrich this context with extra-textual information. One such attempt are factored language models~\cite{Bilmes:2003}. Factored language models represent each word as a factor which can consist of a set of user-defined values, including the word string itself the word's stem, its part of speech or derived prosodic features (among many others). 

Prosodic features have been previously incorporated in factored language models. \newcite{Huang:2007} show that prosodic context in addition to word context improves perplexity of language models conversational data from meetings. 

Here, we integrate prosodic features in language models over child-directed speech. We hypothesize that prosodic features will add complementary disambiguating information to a lexical context and consequently lead to decreased perplexity over unigram or purely lexical n-gram models.

Factored language models allow to flexibly specified relevant context, and generalized backoff methodologies have been developed which allow to choose which conditioning factors to drop and keep at each backoff level of the model. The SRILM language modeling toolkit~\cite{Stolcke:2002} offers an interface for defining such factored language models and backoff strategies, and we implement all our models in this framework.

\subsubsection{Representation of Prosody Features}
N-gram language models are Markov models over a discrete state space. As such all factors must be represented symbolically. Following prior work~\cite{Huang:2007}, we transform our 88-dimensional eGemaps prosody feature vectors into symbolic representations using vector quantization. We take as input the original eGemaps features for each word in our corpus and perform k-means clustering on this set. We then map each word's feature representation to its closest centroid, turning the vector into a symbolic label $\in\{1...k$\}. We experiment with different $k$ below.

\begin{table}
\begin{center}
\begin{small}
 \begin{tabular}{l||ll|ll|ll}
  & $w_{-2}$ & $p_{-2}$ &  $w_{-1}$ & $p_{-1}$ & $\mathbf{w_0}$ & $p_{0}$  \\
  model & look&$k45$ & a&$k22$ & {\bf ball}&$k47$  \\\hline
  uni        &&&&\\
  bi         &&& $\checkmark$ &\\
  tri        & $\checkmark$ && $\checkmark$&&\\
  bi_prosUni &&& $\checkmark$ & &&$\checkmark$\\
  bi_prosBi  &&& $\checkmark$ & $\checkmark$&&$\checkmark$\\
  tri_prosUni&$\checkmark$&& $\checkmark$ & &&$\checkmark$\\
  tri_prosBi &$\checkmark$&& $\checkmark$ & $\checkmark$&&$\checkmark$\\
 \end{tabular}
 \end{small}

 \end{center}
 \caption{Illustration of conditioning context information used in our set of seven language models. The currently predicted word is highlighted in bold face. All words are augmented with prosodic information (as symbols). Models 1--3 are standard lexical n-gram language models. Rows 4--7 augment the bigram and trigram models with prosodic information of the currently predicted word and the immediately preceding word.}
 \label{language-models}
\end{table}

\subsection{Experiments and Results}
We implement seven different language models conditioning on different contexts as illustrated with an example in Table~\ref{language-models}. As a baseline we implement a unigram model ($uni$) with no contextual information. We compare the baseline to models conditioning on increasing lexical context, i.e., a bigram model conditioning on the immediately preceding lexeme ($bi$), and a trigram model considering the two preceding lexemes ($tri$). Finally, we add prosodic factors to the bigram and trigram models in form of either only the currently predicted word's prosodic class ($prosUni$) or both the currently predicted and the immediately preceding words' prosodic class ($prosBi$). Table~\ref{language-models} illustrates our model with a text example.

We experiment with $k \in \{50,100,500\}$ target clusters for prosody vector quantization in order to investigate the dependence of our model on this parameter. 

We train language models on the Brent corpus (106,647 sentences), the Providence corpus (134,690 sentences), and a combination of both corpora (267,337 sentences).\footnote{Number of sentences don't add up because we remove 26,000 sentences from each corpus (13,000 for dev / test each) and do this after concatenating the full Brent and Providence for the combined corpus.} After estimating our models on a training corpus (using SRILM) we report both training set perplexity scores, as well as perplexity scores on a test corpus of 13,000 utterances sampled from both Brent and Providence. This test set remains the same across all training corpora and models. 

Table~\ref{lm-results-testset} presents results when evaluating perplexity on a held-out test set of 13,000 utterances, while Table \ref{lm-results} summarizes training set perplexities of different language models. Unsurprisingly, the baseline unigram model receives highest perplexity scores. The purely lexical bigram and trigram models significantly improve perplexity. 

Training set perplexity consistently improves with richer contexts both in terms of lexical and prosodic information. Adding prosodic features reliably improves over their non-prosodic counterpart (e.g., $tri\_prosUni$ and $tri\_prosBi$ outperform the $tri$ model). Except for the combined corpus, including prosodic bigrams improves performance over prosodic unigrams.

Results are less consistent on test set perplexity. Adding prosodic features further improves perplexity over its non-prosodic counterpart in the case of the trigram model for the Providence corpus and the combined corpora. The $tri\_prosUni$ models, combining a trigram language model with target word prosodic features improves language model performance. The bigram models do not benefit from additional prosodic information. 

In addition, we do not observe an analogous improvement for the Brent corpus. This might be an artifact of the sampling of our test data set which was sampled from the combined corpus where Providence dominates such that the model trained on Brent is effectively tested on `out of domain data'.

% Note that the exact numbers are not comparable {\it between} our two corpora due to the different test sets. 
% The general order of model performance, however, remains stable across corpora suggesting that the results are stable.

\begin{table*}[ht!]
%  \begin{tabular}{lccc}
% \hline
%           & Brent    & Providence & Both\\\hline
% uni   & 173.2115 & 303.2884   & 264.541\\
% bi        & 45.51966 & 82.98096   & 69.74132\\
% tri      & 35.29567 & 65.60900   & 53.39637\\
% bi_prosUni      & 58.08078 & 98.32388   & 87.28918\\
% tri_prosUni    & {\bf 32.71148} & {\bf 62.42590}   & {\bf 50.56628}\\
% bi_prosBi    & 58.13121 & 98.46825   & 87.35312\\
% tri_prosBi  & 35.21605 & 67.48590   & 54.01674\\\hline
%  \end{tabular}
\begin{center}
 \begin{tabular}{l|lll|lll|lll}
 \hline
& \multicolumn{3}{c|}{$k=50$} & \multicolumn{3}{c}{$k=100$} & \multicolumn{3}{|c}{$k=500$}\\
      & Brent & Provid & both & Brent & Provid & Both & Brent & Provid & Both\\\hline
uni &244.75&265.73&263.99&244.75&265.73&263.99&244.75&265.73&263.99\\
bi &76.36&63.48&69.31&76.36&63.48&69.31&76.36&63.48&69.31\\
tri &{\bf 61.50}&46.56&52.59&{\bf 61.50}&46.56&52.59&{\bf 61.50}&46.56&52.59\\
bi_prosUni &130.07&118.31&83.44&129.52&120.91&83.89&125.04&112.78&86.37\\
tri_prosUni &69.21&{\bf 42.00}&{\bf 48.89}&68.37&{\bf 42.64}&{\bf 48.87}&67.82&{\bf 40.81}&{\bf 49.26}\\
bi_prosBi &131.67&119.31&85.15&129.54&121.19&84.84&125.49&113.16&86.57\\
tri_prosBi &71.96&44.21&51.53&70.47&45.31&51.66&69.80&43.24&52.85\\\hline
\end{tabular}
\end{center}
 \caption{{\bf Test set perplexity scores} (13,000 utterances) of our seven models. We compare performance on three corpora (Brent, Providence and both), as well as across different numbers of prosody feature clusters obtained through feature quantization ($k=50,100,500$).}
\label{lm-results-testset}
\end{table*}

\begin{table*}
\begin{center}
\begin{tabular}{l|lll|lll|lll}
\hline
&\multicolumn{3}{c|}{$k=50$}&\multicolumn{3}{c}{$k=100$}&\multicolumn{3}{|c}{k=500}\\
&Brent&Provi&Both&Brent&Provi&Both&Brent&Provi&Both\\\hline
uni&179.50&310.51&266.59&179.50&310.51&266.59&179.50&310.51&266.59\\
bi&35.75&57.67&53.56&35.75&57.67&53.56&35.75&57.67&53.56\\
tri&24.10&37.77&34.62&24.10&37.77&34.62&24.10&37.77&34.62\\
bi_prosUni&26.49&34.84&39.20&23.72&29.92&34.59&18.24&22.04&25.70\\
tri_prosUni&9.78&11.30&12.45&9.09&10.44&11.45&{\bf 7.84}&{\bf 9.23}&9.81\\
bi_prosBi&17.48&19.48&23.45&15.29&16.23&19.55&12.51&13.50&15.36\\
tri_prosBi&{\bf 8.55}&{\bf 9.75}&{\bf 10.27}&{\bf 8.18}&{\bf 9.57}&{\bf 9.80}&7.92&9.74&{\bf 9.62}\\\hline
\end{tabular}
\end{center}
 \caption{{\bf Training set perplexity scores} of our seven language models. We compare performance on three corpora (Brent, Providence and both), as well as across different numbers of prosody feature clusters obtained through feature quantization ($k=50,100,500$).}
 \label{lm-results}
\end{table*}

\subsection{Discussion}
Adding prosodic context to textual context in n-gram language models trained on CHILDES data consistently improves training set perplexity scores, however, did not reveal consistent improvements in test set perplexity. Our experimental setup is influenced by several factors. 

First, we use vector quantization to convert real-valued prosodic feature vectors into prosodic symbols. This process likely leads to substantial loss of information. Secondly, vector quantization involves k-mean clustering and consequently to select a parameter $k$. We test three values, $k \in \{50,100,500\}$, and observed very similar patterns in the results suggesting that the results do not strongly depend on this parameter.

We performed vector quantization on the full 88-dimensional prosodic feature vector. It is possible that a subset of those features is most informative and could be quantized with less of an information loss.

% Moving on to test set perplexity change under added prosodic features (Table~\ref{lm-results-testset}) we can also observe improvements. The $tri\_prosUni$ models, combining a trigram language model with target word prosodic features seems to improves language model performance with the exception for the Brent corpus (which as discussed is not well represented in our held-out test data sample). Compared to training set perplexity improvements, however, results on the test set are less consistent. This suggests that prosodic features do not generalize (as well as lexical features) across data sets. 

Training set perplexity improved consistently and significantly with added prosodic features. We also observed improvements on test set perplexity, however, the trends were less consistent. This suggests that prosodic features may not generalize (as well as lexical features) across data sets.

In the cognitive context of the current investigation we believe that the training set results are still revealing. Our overall results suggest a clear trend of prosodic features being able to improve language model perplexity when trained on child-directed data.

In recent years neural network language models have been introduced and applied to various tasks with great success, largely replacing n-gram language models. In addition to their higher predictive power, neural language models are more amenable to real-valued input vectors than n-gram language models. Few works exist which combine prosody and lexical information in neural network language models~(but see \newcite{Reddy:2015}). We plan to pursue this direction in future work.

\section{Experiment 3: Language-model Representations as Predictors of Age of Acquisition}
The overall motivation behind our work is to investigate the influence of prosodic features in child-directed speech on age of acquisition of words. To this end, in experiment 2 we trained language models over child-directed language containing both lexical and prosodic information. In this experiment we view those language models as (simplistic) models of word learning and investigate the representation learnt by such models as predictors of age of acquisition.

For each word in the age of acquisition data set we approximate the extent to which a language model has learnt a specific word after training. We quantify this extent by the average probability assigned by the model to the word in either the training data or a test data set. We use this averaged probability as a predictor of age of acquisition. We test the same set of language models conditioning on varying lexical and prosodic context as in Experiment 2, as listed in Table~\ref{language-models}.

\subsection{Experiments and Results}
Like in experiment 1 we fit ridge regression models to predict age of acquisition in months for a target set of 600 words. We test as a predictor word frequency as well as word probabilities as estimated by each of our language models (cf.,~Table~\ref{language-models}) and various combinations thereof. We scale all predictors to zero mean and unit variance and map them into log space. Scores are again reported in terms of mean squared error (lower is better). As before, we train and test our models using 10-fold cross validation with random 90\% train / 10\% test splits of the data on the word level. 

\begin{figure*}[ht!]
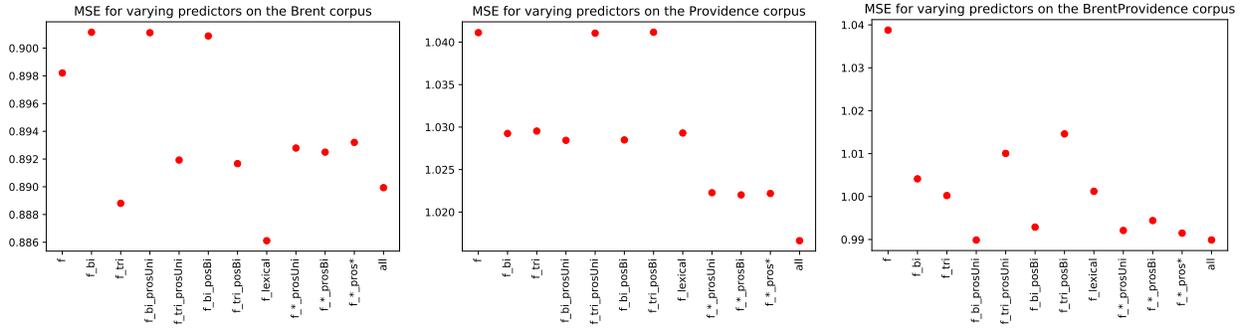

 \includegraphics[width=0.33\textwidth]{\detokenize{./Brent500_srilm_features}}
 \includegraphics[width=0.33\textwidth]{\detokenize{./Providence500_srilm_features}}
 \includegraphics[width=0.33\textwidth]{\detokenize{./BrentProvidence500_srilm_features}}
 \caption{Mean Squared Error (MSE) of various predictors derived from factored language model {\bf test set word probabilities} on our three training corpora (Brent, Providence, both).  Asterisks in the model names indicate that all versions of context types are used (e.g., f_*_prosUni uses as predictors representation from both purely lexical language models, i.e., unigram and bigram).}
 \label{exp-3}
\end{figure*}

\begin{figure*}[ht!]
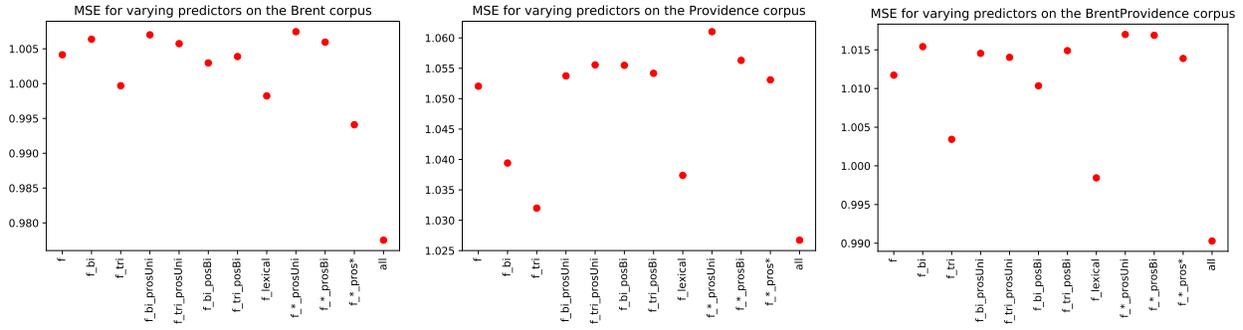

 \includegraphics[width=0.33\textwidth]{\detokenize{./Brent500_test_on_train_srilm_features}}
 \includegraphics[width=0.33\textwidth]{\detokenize{./Providence500_test_on_train_srilm_features}}
 \includegraphics[width=0.33\textwidth]{\detokenize{./BrentProvidence500_test_on_train_srilm_features}}
 \caption{Mean Squared Error (MSE) of various predictors derived from factored language model {\bf training set word probabilities} on our three training corpora (Brent, Providence, both).  Asterisks in the model names indicate that all versions of context types are used (e.g., f_*_prosUni uses as predictors representation from both purely lexical language models, i.e., unigram and bigram).}
 \label{exp-3-test-on-train}
\end{figure*}

Figure~\ref{exp-3} displays results when using language model-induced word probabilities derived from a test set. Figure~\ref{exp-3-test-on-train} summarizes results using language model-induced word probabilities derived from the training set. Models are listed along the x-axis. We combine the word frequency predictor (f) with word representations derived from different our six language models of varying context. Asterisks in the model names indicate that all versions of context types are used (e.g., f_*_prosUni uses as predictors representation from both purely lexical language models, i.e., unigram and bigram).

First we observe that only the test set-derived word probabilities (Figure~\ref{exp-3}) reliably explain variance in the age of acquisition data beyond the frequency detector. Training set derived probabilities (Figure~\ref{exp-3-test-on-train}) are not informative in this experiment. Below, we focus on the test set probability results in Figure~\ref{exp-3}.

Throughout all corpora adding language model features improve mean squared error compared to using frequency as the sole predictor. This is encouraging, demonstrating that the induced language model word probabilities explain variance in the age of acquisition data beyond the variance explained by the already strong predictor of word frequency. 

The contribution of prosodic information is less clear. Patterns vary across corpora. For the Providence corpus (center) and the combined corpora (right) language models including prosodic features can lead to reduced mean squared error compared to the purely lexical models $f\_bi$ and $f\_tri$. Finally, combining all predictors (rightmost point; all) consistently leads to best MSE with the exception of the test set perplexity for the Brent corpus.

\subsection{Discussion}
The language model derived predictors in form of word type probabilities as averaged word token probabilities in the test data set strongly depend on the characteristics of this test data set. In particular, we expect estimates of infrequent words to highly vary based on their frequency and context of occurrence test set.

As noted in the discussion of experiment 2 the test set from which word probabilities were derived is dominated by Providence data. This makes estimation of parameters based on the Brent corpus particularly brittle.

Experiment 3 builds on representations induced by language models on fairly small training corpora. In addition the underlying prosodic information in Experiment 2 was highly transformed and filtered through vector quantization. These inaccuracies accumulate in our experimental setup and are likely a reason for the inconclusive results.

\section{Conclusions}
Early word learning strongly depends on a number of linguistic and extra-linguistic factors. Apart from the content of child-directed language, pragmatic cues such as speech prosody, gaze and gestures play an important role of guiding children's attention towards correct word-meaning mappings. In this work we explored the utility of prosody in raw, large-scale child-directed speech as a predictor of age of acquisition.

The motivation behind our work was two-fold. First, corpora of child-directed data are augmented to a large portion with audio recordings of the interaction. This allowed us to construct a large-scale multi-modal corpus of transcribed speech and its prosodic features. Secondly, prior work on predicting age of acquisition used features which were derived from child language input (e.g., frequency or mean length of utterance), but not the characteristics of the raw input itself. Here we bridge this gap.

We conducted three sets of experiments. First, we used the automatically obtained prosodic features of each word type (as averaged over token occurrences) from raw child-directed speech. We explored these features as predictors of age of acquisition. We were able to identify a subset of those features which seemed particularly useful for this task. Experiments 2 and 3 utilized token-level prosodic information rather than marginalizing it into one type-level representations. In experiment 2, we trained language models with different conditioning contexts on child-directed data. Results showed that conditioning on prosodic information in addition to lexical context can improve language model perplexity. Finally, experiment 3 views the estimated language models as (simplistic) models of language learning and used the induced word representations, in the form of word probabilities assigned by the language models, as predictors in regression models for age of acquisition. 

To the best of our knowledge we present the first large-scale data-driven investigation of prosodic features of child-directed speech on age of acquisition of words. While our initial models are rather simplistic and results mixed, our work opens up various interesting directions for future investigations. First, we would like to incorporate our prosodic features into richer models of word learning. One could for example augment the cross-situational models of \newcite{Fazly:2010} with prosodic features as a more realistic mechanism to construct extra-linguistic scenes.

We also plan to further investigate the utility of prosodic information for language models on child-directed speech. Using n-gram models forced us to discretize our real-valued prosodic feature vectors. More recent neural network language models allow more naturally to incorporate feature vectors, and provide richer methods of using and selecting relevant aspects of such features. Investigating the utility of multi-modal information through neural language models is an exciting future direction to pursue.

\bibliography{bib}
\bibliographystyle{acl2012}

\end{document}